\pdfoutput=1
\documentclass[11pt]{article}

\usepackage{ACL2023}

\usepackage{times}
\usepackage{latexsym}

\usepackage[T1]{fontenc}
\usepackage[utf8]{inputenc}

\usepackage{microtype}

\usepackage{inconsolata}

\usepackage{booktabs}
\usepackage{tabularx}
\usepackage{caption}
\usepackage{graphicx}
\usepackage[dvipsnames]{xcolor}
\usepackage{listings}
\usepackage{floatrow}
\usepackage{newfloat}
\newfloatcommand{capbtabbox}{table}[][\FBwidth]
\newfloat{lstfloat}{htbp}{lop}
\floatname{lstfloat}{Prompt}
\title{Coordinates from Context: \\Using LLMs to Ground Complex Location References}

\author{Tessa Masis \normalfont{\emph{(they/them)}} \quad
    \bf{Brendan O'Connor}
\normalfont{\emph{(he/him)}}
    \\
    University of Massachusetts Amherst, MA, USA\\
    \texttt{\{tmasis, brenocon\}@cs.umass.edu}
  }

\begin{document}
\maketitle
\begin{abstract}
Geocoding is the task of linking a location reference to an actual geographic location and is essential for many downstream analyses of unstructured text. In this paper, we explore the challenging setting of geocoding compositional location references. Building on recent work demonstrating LLMs' abilities to reason over geospatial data, we evaluate LLMs' geospatial knowledge versus reasoning skills relevant to our task. Based on these insights, we propose an LLM-based strategy for geocoding compositional location references. 
We show that our approach improves performance for the task and that a relatively small fine-tuned LLM can achieve comparable performance with much larger off-the-shelf models.\footnote{Our best-performing fine-tuned model is available at \url{https://huggingface.co/tmasis/geocoding-complex-location-references}
}

\end{abstract}

\section{Introduction}

Extracting geospatial information from unstructured text is essential for many downstream analyses such as disaster response \citep{kumar2019location}, disease surveillance \citep{lee2013real}, 
%analyzing language variation \citep{huang2016understanding}, 
and historical event analysis 
\cite{tateosian2017tracking}. Geocoding is the task of linking a location reference to an actual geographic location, usually represented by a set of coordinates or an ID in a geographic database. 

While there is a large body of work on geocoding locations referred to with explicit names (e.g. "in Gaza City"), there has been little attention given to locations referred to with compositional descriptions involving other toponyms (e.g. "about midway between Winsford and Crewe") which occur frequently in domains that do not have names for geographic regions of interest, e.g. environmental impact statements, food security outlooks \cite{laparra2020dataset}. In such a setting, existing geocoding tools are not able to link the location being described by the entire phrase, only locations mentioned within the phrase \cite{zhang2024survey}.  
In addition, most geocoding tools can only link to locations already in a geographic database, which often do not include compositional locations.

\begin{figure}[t]
    \centering
    \includegraphics[width=.96\textwidth]{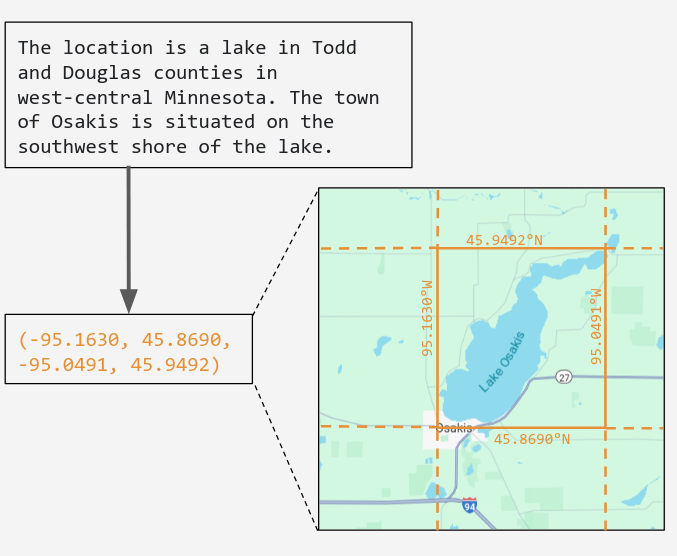}
    \caption{An illustrative example of our task. Given a compositional location description, we expect the model to predict the location's bounding box (defined by two pairs of latitude-longitude coordinates).  } 
    \label{fig:example}
\end{figure}

In this paper, we address the task of geocoding compositional location descriptions (see Figure \ref{fig:example}) and how LLMs can best be leveraged for it.
Our work makes the following contributions:
\begin{itemize}
    \item We provide findings on LLMs' geospatial \emph{knowledge} versus geospatial \emph{reasoning} abilities through task-specific isolated evaluations of each. We find that off-the-shelf LLMs struggle more with generating accurate geospatial knowledge than with reasoning over it.
    \item Based on these insights, we propose the first end-to-end strategy for geocoding compositional location references, leveraging both the geospatial reasoning capabilities of LLMs and the geospatial knowledge bases of traditional geoparsers. Furthermore, our novel use of bounding boxes\footnote{A \emph{bounding box} is a rectangular area defined by two longitudes and two latitudes, and written in the standard format of $\{ \mathrm{lon}_{min}, lat_{min}, lon_{max}, lat_{max} \}$ where $lon \in [-180, 180]$ and $lat \in [-90, 90]$.} to link locations provides a simple and informative way to ground locations that do not exist in a geographic database. 
    \item We thoroughly evaluate components of our approach, demonstrating that it outperforms prior work and that a relatively small fine-tuned LLM can achieve comparable performance with much larger off-the-shelf models.
    
    %\item (possibly) another contribution is making the dataset easier to use? i.e. making bounding boxes (including of mentioned locations) an easily accessible part of the dataset; easier to parse text format

\end{itemize}

\section{Related Work}

\textbf{Geocoding}, also known as toponym resolution, aims to link some reference to a geographic entity to the correct reference in a target database or coordinate system \cite{zhang2024survey}.
The majority of previous work has focused on geocoding location references that have an explicit location name, with difficult settings typically concerning ambiguous location names (e.g. "Paris" may refer to Paris, TX, USA or Paris, FR).
Some researchers have looked at geocoding location references without explicit location names, such as indirect location references (e.g. "the Austrian capital") or nonstandard variants of location names (e.g.\ ``Beantown'' refers to Boston, MA, USA) \cite{dredze2013carmen, delozier2015gazetteer, kulkarni2021multi, masis2024earth}. 

Very little attention has been given to geocoding complex or compositional location references, which do not contain an explicit location name and instead describe the location in relation to other toponyms \cite{al2019towards}. Thus far, there is only one dataset for this setting \cite{laparra2020dataset}. The authors introduce five baseline methods for the task, although none using pretrained language models.

%Unlike traditional geocoding settings, geocoding compositional location references requires higher-level reasoning to understand described relationships between geographic entities and to combine geospatial interpretations of these relationships. Very little work in geocoding has leveraged reason
%with most work learning co-occurrences between tokens and geographic locations. - or laparra which uses semantic parsers 

Compositional location references must be linked to a coordinate system, not to a geographical database (e.g.\ GeoNames\footnote{\url{https://www.geonames.org}} or
OpenStreetMap\footnote{\url{https://www.openstreetmap.org}}), since compositional locations often do not exist in geographic databases. While some prior work has examined geocoding without such databases \cite{delozier2015gazetteer, kulkarni2021multi, sharma2024spatially}, they typically link locations to point coordinates which do not represent large or compositional geographic entities well. In our work, we link locations to a bounding box, which defines an area in terms of four sets of coordinates and allows us to ground locations much more precisely.

\textbf{Geoparsing} is the broader task of both identifying and linking location references in unstructured text \citep{wang2019enhancing}, essentially combining toponym recognition \cite{hu2023location} and geocoding. 
%Popular English geoparsers include CLIFF-CLAVIN \cite{d2014cliff}, CamCoder \cite{gritta2018melbourne}, and the Edinburgh geoparser \cite{grover2010use}. 
As with geocoding, no geoparsing work to date has addressed compositional location references and methods primarily link location references to a geographic database or a single set of coordinates.

\textbf{LLMs' geospatial knowledge and reasoning skills} have begun to be evaluated in a growing body of work. (We define \emph{geospatial}, following the ISO's definition\footnote{\url{https://isotc211.geolexica.org/concepts/202/}}, as an 'implicit or explicit reference to a location relative to the Earth' typically involving coordinates or addresses.)
Regarding geospatial knowledge, 
%Building on earlier work examining language models' geospatial knowledge representations, 
recent work has suggested that LLMs learn and use an internal model of space which encodes locations' approximate real-world geo-coordinates \cite{gurnee2024language, chen2023more}.
Related studies have examined the disparities and differences in LLMs' geospatial knowledge due to factors such as language used \cite{faisal2023geographic, li2024land} or model size \cite{godey2024scaling, bhandari2023large}.
To help explain how LLMs may acquire such knowledge, an examination of the Common Crawl Corpus -- often used in the pretraining of LLMs -- found that 19\% of documents contained geospatial data and that a prominent source was Google Maps URLs, which frequently contain both coordinates and a natural language location name \cite{ilyankou2024quantifying}. 
However, to our knowledge no work has yet empirically evaluated LLMs as geospatial knowledge bases for downstream tasks.

Regarding geospatial reasoning, recent work evaluating off-the-shelf LLMs has emphasized their promise but also their limitations for geospatial reasoning tasks \cite{ji2025foundation}. Researchers have found that LLMs such as GPT-4o and Claude 3 Sonnet have limited success with reasoning over geospatial data, struggling to translate qualitative natural language terms to geospatial contexts \cite{osullivan2024metric} and performing poorly at tasks like simple route planning \cite{xu2024evaluating}. 
Importantly, these studies show that LLMs have the capability to reason over geospatial information in both natural language and geo-coordinate formats. 

Thus far, no work has directly compared LLMs' geospatial knowledge versus reasoning performances. Furthermore, no work has explored LLMs' geospatial knowledge of or ability to reason over geo-entities defined by bounding boxes.

\textbf{Using LLMs for geoparsing} has recently been explored, with methods primarily using LLMs for the toponym recognition step \cite{hu2024toponym, harrod2024text, xu2024evaluating, hu2023geo}, to generate training data \cite{yan2024georeasoner}, or to embed natural language descriptions of point coordinate locations \cite{he2025geolocation}. 
Thus far, LLMs have not been used for generating geospatial information or for reasoning-heavy settings.

%LLMs recalling non-geo factual information (e.g. population) about specific locations: can do so well when augmented with OSM data (Manvi et al. 2024a), geographic disparities (Moayeri et al. 2024, Manvi et al. 2024b)

\section{Disentangling LLMs' Geospatial Knowledge versus Reasoning}
\label{sec:disentangle}

\begin{figure}[t]
    \centering
    \includegraphics[width=.96\textwidth]{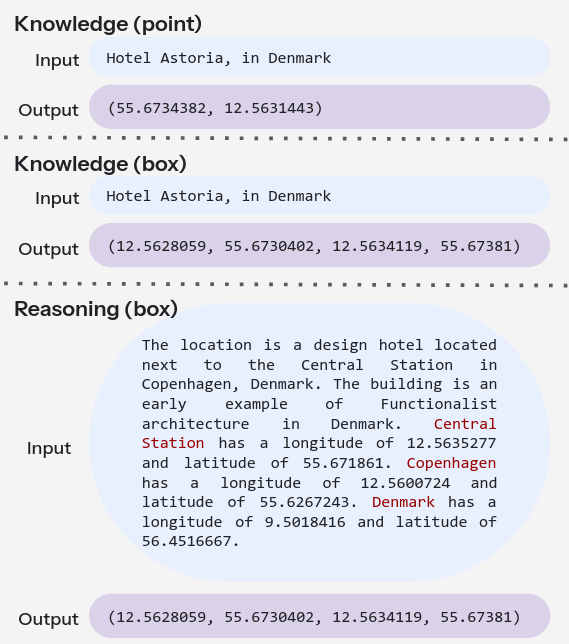}
    \caption{Illustrative examples of our LLM geospatial knowledge and reasoning baselines; all examples refer to the same location. Knowledge baseline: the LLM is given only the location's name. Reasoning baseline: the LLM is given only a description of the location and mentioned locations' coordinates (mentioned location names in red, for emphasis).
    Point refers to a center coordinate prediction (written as $\{ lat, lon \}$); box refers to a bounding box prediction (written as $\{ lon_{min}, lat_{min}, lon_{max}, lat_{max} \}$).
    } 
    \label{fig:disentangle}
\end{figure}

Here, we briefly evaluate LLMs for their geospatial knowledge generation versus geospatial reasoning abilities (see Figure \ref{fig:disentangle}).
These two capabilities -- that is, generating relevant factual knowledge and reasoning over provided factual knowledge -- 
are necessary for the task of geocoding compositional locations (see task definition in \S\ref{sec:task-definition}, below), so isolated evaluations of each will help us understand how to best leverage LLMs for this task.
%For the former, we wish to only evaluate the model's ability to generate relevant factual knowledge; for the latter, we wish to only evaluate the model's ability to reason over provided factual knowledge. 

Accordingly, for our geospatial knowledge baseline we prompt the LLM to generate a structured geospatial representation (either the center coordinate or the bounding box) of a location \emph{identified unambiguously by name} (Fig. \ref{fig:disentangle}, top half). 
Previous work only examined LLMs' knowledge of geo-entities represented as center coordinates -- here, we include both center coordinate and bounding box representations, allowing us to directly compare LLMs' knowledge of each.
We include both the location's name and country in the prompt, since a name can refer to multiple locations (e.g. "San Jose" is a city in both Costa Rica and the USA) and including country information reduces reasoning needed to resolve such ambiguities. 

For our geospatial reasoning baseline, we prompt the LLM to generate the bounding box of an \emph{unidentified location described in relation to other geo-entities} (Fig. \ref{fig:disentangle}, bottom half). These prompts include a description of the location and the center coordinates of locations mentioned in the description. The described location's name is not included so that the only way to predict its bounding box is to reason over the given information.

\subsection{Experimental Setup}
\label{sec:experiment-setup}

\textbf{Data.} We use the \textsc{GeoCoDe} dataset\footnote{\url{https://github.com/EgoLaparra/geocode-data}}, which contains location descriptions from English Wikipedia articles that are linked to the OpenStreetMap (OSM) geodatabase \cite{laparra2020dataset}. 
For the current evaluation, we use \textsc{GeoCoDe}'s test set of 1,000 manually curated compositional location descriptions. Created by randomly sampling Wikipedia pages, the dataset includes locations from all continents.

\textbf{Representing locations.} By leveraging the OSM links in \textsc{GeoCoDe}, we retrieve gold geospatial information to represent each described location and location mentioned in a description. 
%This allows us to represent locations with center coordinates, bounding boxes, or polygons. 
Locations can be represented as center coordinates, bounding boxes, or polygons, each with its own pros and cons. 
\emph{Center coordinates}, which represent a location as a point, are used by the vast majority of geocoding datasets and are the simplest way to ground a location that does not exist in a geographic database. While adequate for point locations or small landmarks, they fall short when representing larger entities such as states or countries. 
\emph{Polygons} represent locations with an unlimited sequence of coordinates defining a closed shape, so are much more complex and thus rarely used; at present, there are no practical end-to-end geocoding systems using polygons. 
\emph{Bounding boxes} represent a location as a rectangle-shaped area via two latitude-longitude coordinate pairs, efficiently encoding basic size and shape.

In this work, we primarily represent locations with bounding boxes. 
While not used before for geocoding, we argue that bounding boxes offer a happy medium between center coordinates and polygons -- they can encode areal information about the location while still being simple enough that off-the-shelf LLMs can reason over them.

\textbf{Metrics.} We use five metrics to evaluate predictions. Two are common geocoding metrics: \emph{coverage} is the percentage of examples where the model produces a prediction, and \emph{distance error} is the distance between the predicted and gold center coordinates or bounding box centroids. The other three metrics, introduced in \citeauthor{laparra2020dataset}, measure the overlapping area between a predicted and gold bounding box: \emph{area precision} is the fraction of overlapping area out of the entire predicted area, \emph{area recall} is the fraction of overlapping area out of the entire gold area, and \emph{area F1-score} is the harmonic mean of the precision and recall (calculated at test set-level, following prior work).
%(following the standard definition of F1-score). 
Since these metrics measure area overlap, they are not used to evaluate center coordinate predictions.

\textbf{Models.} We evaluate three off-the-shelf open-source LLMs: Qwen2.5-72B-Instruct, Llama3.3-70B-Instruct, and Llama3.1-405B-Instruct. The models are evaluated in a few-shot manner, with two examples included in the system prompt (see Appendix \ref{sec:appendix-prompts} for full prompts used).

\subsection{Results}

Table \ref{tab:disentangle} reports the results of our geospatial knowledge versus reasoning baselines (also see Table \ref{tab:disentangle-complete} in Appendix). We observe that all models are decidedly better at the geospatial reasoning baseline than the geospatial knowledge one, with lower average distance errors by more than 140km and consistent increases in area F1-score. The Qwen 72B model appears to have the biggest gap between knowledge and reasoning, with a difference of 300km between distance errors and of .164 between F1-scores. 
Additionally, increasing model size appears to improve geospatial knowledge more than geospatial reasoning (e.g. F1-score for Llama 70B versus 405B is .137 versus .195 for bounding box knowledge, but .239 versus .235 for reasoning). 
Finally, we note that all models have slightly better performances when predicting a point than a bounding box for the knowledge baseline, with lower distance errors of 10 to 50km.

\begin{table}[t]
\centering
  \begin{tabular}{llrr}
    \toprule
    \textbf{Experiment} & \textbf{LLM} & \textbf{Dist. $\downarrow$} & \textbf{F1 $\uparrow$}\\
    \midrule
    Knowledge & Qwen 72B & 326.1  & -- \\
    (point) & Llama 70B &215.4 &  --\\
    & Llama 405B &  216.6 &  --\\
    \midrule
    Knowledge  & Qwen 72B &  333.3 &  .087 \\
    (box)& Llama 70B & 221.8 &  .137 \\
    & Llama 405B & 263.8 & .195 \\
    \midrule
    Reasoning  & Qwen 72B & \underline{\textbf{36.2}} & \underline{\textbf{.251}} \\
    (box) & Llama 70B &  \underline{82.9}  & \underline{.239} \\
    & Llama 405B & 92.4  & .235 \\
    \bottomrule
  \end{tabular}
  \caption{Evaluations from our geospatial knowledge and reasoning baselines, with average distance error (km) and area F1-scores. Point refers to a center coordinate prediction; box refers to a bounding box prediction.
  The best scores are bolded, the top 2 are underlined.  } 
  \label{tab:disentangle}
\end{table}

While previous work has argued that LLMs have some amount of geospatial knowledge, the evidence has mostly demonstrated learned connections between location names (usually cities) and their center coordinates \cite{bhandari2023large, gurnee2024language}.
We explore LLMs' geospatial knowledge with both center coordinates and bounding boxes, a more complex geo-entity representation, and our results caution against using LLMs as reliable geospatial knowledge bases in this setting.
While the LLMs appear to perform slightly better when predicting a point than a bounding box, which may be due to point representations of geo-entities being a more common form of geospatial data in LLM pretraining data \cite{ilyankou2024quantifying}, both settings have relatively high average distance errors of over 200km.

On the other hand, our results show that LLMs perform promisingly on our geospatial reasoning baseline, with average distance errors of less than 35km. This supports prior work demonstrating that LLMs can reason over geospatial information encoded as both natural language and complex geometries \cite{ji2025foundation}, even if complex geometries are not frequently in their pretraining data.

\section{Geocoding Complex Location References}
\label{sec:geocode}

The above analysis reveals how LLMs can reason over geospatial information in both natural language and coordinate-based formats, but struggle with accurately generating such information. This motivates us to augment LLMs' geospatial knowledge in our proposed geocoding method. In this section, we will first formally define our task and then explain our proposed and baseline methods.

\subsection{Task Definition}
\label{sec:task-definition}

We formally define our task of geocoding compositional descriptions as follows. 
Each input to the geocoding system is a text description $d_l$ of a location $l$, where the set of locations mentioned in description $d_l$ is referred to as $M_l = \{m_1, m_2, ..., m_{|M_l|}\}$. The description $d_l$ does not include the name of location $l$.

\begin{figure}[t]
    \centering
    \includegraphics[width=.96\textwidth]{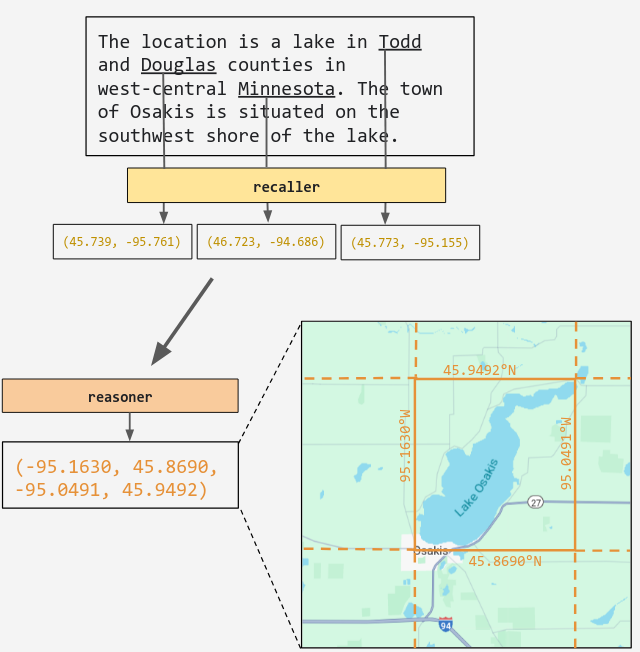}
    \caption{We define our geocoding task as a combination of two functions. The first function (the \texttt{recaller}) retrieves mentioned locations' geographical information, e.g. center coordinates. The second function (the \texttt{reasoner}) uses both the location description and mentioned location information to generate the described location's bounding box. } 
    \label{fig:task-definition-example}
\end{figure}

The geocoding system is a combination of two functions (Figure \ref{fig:task-definition-example}). The first,
$$g(m_i) = q_i$$ 
maps a mentioned location $m_i$ to corresponding geographical information $q_i$, such as the mentioned location's center coordinates. We refer to $g$ as the \texttt{recaller}. 
%The function $g$ can be an oracle retriever that always returns correct geographical information, an LLM which may generate incorrect geographical information, or there may be no system at all and no geographical information is returned. 
The second function
$$ f(d_l, M_l, g) = b_l$$ 
maps the location description and information about locations it mentions to the minimum bounding box $b_l$ enclosing the location $l$. We refer to $f$ as the \texttt{reasoner}.

\subsection{Our Approach}

In our proposed approach, we use a traditional geoparsing tool to augment an LLM's geospatial knowledge (we refer to our approach as \underline{Geoparser-augmented}). 
We first use a traditional geoparsing tool as the \texttt{recaller}, mapping any location mentioned in the compositional description to a set of latitude and longitude coordinates. We then use an LLM as the \texttt{reasoner}, which uses the location description and mentioned locations' coordinates to generate the described location's bounding box.\footnote{The \texttt{reasoner} for the Geoparser-augmented approach has the same expected inputs and outputs as the reasoning baseline in \S\ref{sec:disentangle}.} We evaluate both off-the-shelf and fine-tuned LLMs; see \S\ref{sec:training} for training details.

By augmenting with a traditional geoparser, we are able to add structured information from an external knowledge source to the LLM's current context. This approach is related to the broader field of tool-augmented LLMs, in which LLMs leverage external tools \cite{mialon2023augmented}. 
While traditional geoparsing tools cannot reason over compositional location descriptions or provide complex geospatial information (such as bounding boxes), they can link mentioned locations to a pair of coordinates. The mentioned locations' coordinates can then serve as an accurate geospatially grounded starting point for the LLM to reason over where the described location is in relation to them.

In our experiments, we first simulate an oracle geoparser by retrieving locations' center coordinates from the OSM database.
This allows us to focus on the variation in performance introduced by LLMs, not by third-party geoparsing tools. 
%After identifying the best performing LLM in this setting, we then evaluate it with a real geoparser. 
We then evaluate a more realistic setup with a real geoparser as the \texttt{recaller}.
%and a fine-tuned LLM as the \texttt{reasoner}.
For this, we use the Google Maps API\footnote{\url{https://developers.google.com/maps/documentation/geocoding}} which offers a popular geoparsing service broadly recognized as reliable \cite{ozer2020creating}.

\begin{table*}[t]
\centering
  \begin{tabular}{llrrrr}
    \toprule
    \textbf{Approach} & \texttt{reasoner} & \textbf{Distance (km) $\downarrow$} & \textbf{AreaPrec $\uparrow$} & \textbf{AreaRec $\uparrow$} & \textbf{AreaF1 $\uparrow$}\\
    \midrule\midrule
    Direct & Llama 8B &  4427.8 & .000 & .000 & .000 \\
    **Geo.-aug. & Llama 8B &  7658.5 & .006 & .037 & .010 \\
    \midrule
    Direct & Llama 70B &  120.0 & .200 & .171 & .184 \\
    **Geo.-aug.. & Llama 70B &  82.9 & .166 & \underline{\textbf{.428}} & \underline{.239} \\
    \midrule
    Direct & Llama 405B &  \underline{69.0} & \underline{\textbf{.305}} & .205 & \underline{\textbf{.245}} \\
    **Geo.-aug. & Llama 405B & 92.4 & .164 & \underline{.413} & .235 \\
    \midrule
    Direct & FT Llama 8B &  88.9 & .132 & .094 & .110 \\
    **Geo.-aug. & FT Llama 8B & \underline{\textbf{28.0}} & \underline{.237} & .186 & .208\\
    Geo.-aug. & FT Llama 8B & 152.2 & .231 & .170 & .196 \\
    \bottomrule
  \end{tabular}
  \caption{Results comparing Geoparser-augmented versus Direct approaches for Llama models on \textsc{GeoCoDe} dataset, with average distance error, area precision, area recall, and area F1-scores. FT refers to a fine-tuned model; * denotes an oracle geoparser. The best scores are bolded, the top 2 are underlined.  }
  \label{tab:geoparser-vs-direct-llama}
\end{table*}

\begin{table*}[t]
\centering
  \begin{tabular}{llrrrr}
    \toprule
    \textbf{Approach} & \texttt{reasoner} & \textbf{Distance (km) $\downarrow$} & \textbf{AreaPrec $\uparrow$} & \textbf{AreaRec $\uparrow$} & \textbf{AreaF1 $\uparrow$}\\
    \midrule\midrule
    Direct & Qwen 14B & 1156.4 & .058 & .050 & .054 \\
    **Geo.-aug. & Qwen 14B & 152.1 & .134 & \underline{.498} & .211 \\
    \midrule
    Direct & Qwen 72B  & \underline{114.7} & .113 & .162 & .133 \\  
    **Geo.-aug. & Qwen 72B &  \underline{\textbf{36.2}} & \underline{.191} & .368 & \underline{.251} \\
    \midrule
    Direct & FT Qwen 14B & 1944.1 & .037 & .122 & .057 \\
    **Geo.-aug. & FT Qwen 14B &  135.9 & .165 & \underline{\textbf{.499}} & .248 \\
    Geo.-aug. & FT Qwen 14B  & 240.9 & \underline{\textbf{.203}} & .384 & \underline{\textbf{.266}} \\
    \bottomrule
  \end{tabular}
  \caption{Results comparing Geoparser-augmented versus Direct approaches for Qwen models. FT refers to a fine-tuned model; * denotes an oracle geoparser. The best scores are bolded; the top 2 are underlined. }
  \label{tab:geoparser-vs-direct-qwen}
\end{table*}

\subsection{Baselines}

We evaluate several variations of our method as baselines. 

\underline{Direct}: we use an LLM for the \texttt{reasoner} but no explicit system for the \texttt{recaller}. In other words, an LLM must reason over just the natural language location description to generate the described location's bounding box. 
As with our main approach, we evaluate both off-the-shelf and fine-tuned LLMs for the \texttt{reasoner} (see \S\ref{sec:training} for details).

\underline{End-to-end LLM}: we use LLMs for both the \texttt{recaller} and the \texttt{reasoner}. 
We first prompt the \texttt{recaller} to generate point coordinates for the mentioned locations; for this, we use off-the-shelf Llama 70B since it had the best performance in our geospatial knowledge baseline in \S\ref{sec:disentangle}. 
We then prompt the \texttt{reasoner} to use both the location description and LLM-generated point coordinates to generate the described location's bounding box. This is the same as the \texttt{reasoner} for the Geoparser-augmented approach, 
so we use the best performing fine-tuned model identified in that setting.
This LLM-only pipeline offers a more convenient alternative to the Geoparser-augmented approach.

We also include results from the best performing method in prior work. \underline{GBSP}: in this method, the \texttt{reasoner} is a grammar-based semantic parser and uses polygons of mentioned locations \cite{laparra2020dataset}. 
We note that this method is not a practical end-to-end system since there is no \texttt{recaller}, and currently no system exists which can return polygons of mentioned locations.
%Since no systems currently exist which can return polygons of mentioned locations, the authors simulate such a \texttt{recaller} by retrieving locations' polygons from the OSM database.

\begin{table*}[t]
\centering
  \begin{tabular}{llrrr}
    \toprule
    \textbf{Approach} & \texttt{reasoner} & \textbf{Coverage (\%) $\uparrow$} & \textbf{Distance (km) $\downarrow$} & \textbf{AreaF1 $\uparrow$}\\
    \midrule
    Geo.-aug. & FT Qwen 14B & 90.8 & \textbf{240.9} & \textbf{.266} \\
    End-to-end LLM & FT Qwen 14B & \textbf{95.9} & 627.5  & .167 \\
    GBSP & Semantic parser & 52.8 & -- & .240\\
    \bottomrule
  \end{tabular}
  \caption{Results comparing Geoparser-augmented, End-to-end LLM, and GBSP. The best scores are bolded.}
  \label{tab:extended-results}
\end{table*}

\section{Experiments}

\subsection{Experimental setup}
\label{sec:training}

\textbf{Data.} We use the \textsc{GeoCoDe} dataset, which is discussed in \S\ref{sec:experiment-setup}. We use its manually curated test set as our test set. 
\textsc{GeoCoDe} also has a train set of over 350K examples, which may contain simpler descriptions (i.e. not compositional) or broken OSM links. For our supervised fine-tuning experiments, we use a subset of 13K examples randomly sampled from this train set.

\textbf{Models and training.} We evaluate a range of state-of-the-art open-source models across sizes and from multiple families. For our methods that use an off-the-shelf model, we evaluate Llama3.1-8B-Instruct, Llama3.3-70B-Instruct, Llama3.1-405B-Instruct \cite{dubey2024llama}, Qwen3-14B-Instruct \cite{yang2025qwen3}, and Qwen2.5-72B-Instruct \cite{yang2024qwen2}. For methods that use a fine-tuned model, we evaluate Llama3.1-8B-Instruct and Qwen3-14B-Instruct.\footnote{We evaluated Qwen3 models in the non-thinking mode.} 

For fine-tuning, we conduct supervised fine-tuning and use parameter-efficient fine-tuning via low-rank adaptation \cite{hu2021lora}. 
Separate models are fine-tuned for each approach, with training examples including the appropriate mentioned location geospatial information. 
We fine-tune low-rank adapter matrices for 1 epoch with a learning rate of 2$e^{-4}$ and batch size of 4. 
See Appendix \ref{sec:appendix-prompts} for more details, including all prompts used.

\subsection{Results}
\label{sec:results}

\textbf{Geoparser-augmented and Direct.} We see that the Geoparser-augmented approach generally performs better than the Direct approach (Tables \ref{tab:geoparser-vs-direct-llama} and \ref{tab:geoparser-vs-direct-qwen}; complete tables in Appendix \ref{sec:appendix-tables}). All models, with the exception of Llama 405B, have lower distance errors and higher F1-scores with the Geoparser-augmented approach. 
When using the Google Maps geoparser instead of the oracle geoparser, F1-scores do not change substantially although average distance errors do increase. 
The fine-tuned Qwen 14B model actually has a slightly higher area F1-score when using the Google Maps geoparser, although it also has relatively lower coverage (91\% versus 96\%). 

Interestingly, we find that the Direct approach often results in better area precision than area recall while the opposite is true for the Geoparser-augmented approach, which generally has better area recall than area precision. The former indicates smaller, more precise predicted areas while the latter indicates larger, less precise predicted areas.
This is surprising, since the Geoparser-augmented method provides the LLM with accurate geospatial information about mentioned locations which should hypothetically allow the predicted bounding box to be \emph{more} precise and should not necessarily affect the predicted bounding box's size. 
We note a similar phenomenon in our earlier experiments in \S\ref{sec:disentangle}, where LLMs had better area precision than recall on the knowledge baseline and better area recall than precision on the reasoning baseline (Table \ref{tab:disentangle-complete}). There, too, the method that does not provide mentioned locations' geospatial data has higher area precision while the method that does provide them has higher area recall. 
We further investigate this phenomenon below in \S\ref{sec:qual}.

\textbf{End-to-end LLM and GBSP.}
In Table \ref{tab:extended-results}, we compare the Geoparser-augmented approach with End-to-end LLM and GBSP (complete table in Appendix \ref{sec:appendix-tables}). We observe that the End-to-end LLM approach has relatively poor performance compared to the Geoparser-augmented, with a lower F1-score by .1 and a higher distance error by 390km.
These results support our earlier experiments in cautioning against using LLMs as geospatial knowledge bases. 
We note that End-to-end LLM does perform better than the Direct method using the same model (e.g. F1-scores of .167 versus .057; Table \ref{tab:geoparser-vs-direct-qwen}), indicating that using an LLM to generate mentioned locations' geospatial information is better than providing no information at all.

We also observe that the GBSP method has a relatively high F1-score of .240 but a low coverage, only making predictions for 53\% of the test set. 

\textbf{Comparing LLMs.} Comparing model families, we see a similar pattern as in \S\ref{sec:disentangle} where Qwen models (Table \ref{tab:geoparser-vs-direct-qwen}) perform slightly better when mentioned locations' geospatial information is provided and Llama models (Table \ref{tab:geoparser-vs-direct-llama}) have a slight edge when it is not. This suggests that Llama models may have better geospatial knowledge representations while Qwen models may have better geospatial reasoning abilities, which would support prior work demonstrating Qwen's stronger reasoning abilities \cite{wang2025octothinker}.
However, we note that disparities in performance are more dependent on model size than model family. 

We also observe that while fine-tuned models do not surpass the performance of larger off-the-shelf models, they can achieve comparable performance. We see this most with the Geoparser-augmented approach, where fine-tuning models results in slightly bigger gains than with the Direct approach (e.g. fine-tuning improves Llama 8B's F1-score by .1 more and Qwen 14B's F1-score by .03 more in the Geoparser-augmented setting).

\textbf{Overall.} Our best performance is achieved by Qwen models with the Geoparser-augmented approach -- Qwen 72B has an F1-score of .251 and an average distance error of 36km, and fine-tuned Qwen 14B has an F1-score of .248 with an oracle geoparser and .266 with the Google Maps geoparser. 
Both of these methods outperform GBSP, the best performing method from prior work, despite having access to less geospatial information (i.e. mentioned locations' center coordinates versus polygons). 
We note that the fine-tuned Qwen 14B model achieves comparable performance to much larger off-the-shelf models, with a similar F1-score to Qwen 72B and higher F1-scores than Llama 70B or 405B, although also higher distance errors. 

We offer a few practical suggestions for researchers interested in using LLMs for geocoding complex location references:
\begin{itemize}
    \item For best overall performance or to prioritize area recall or distance errors, we recommend the Geoparser-augmented approach with a smaller fine-tuned LLM (e.g. fine-tuned Llama 8B or Qwen 14B) or a medium off-the-shelf model (e.g. Qwen 72B, Llama 70B). Since we found that scaling model size is less effective for improving geospatial reasoning than knowledge (\S\ref{sec:disentangle}), using the largest available LLM with the Geoparser-augmented approach is likely unnecessary. 
    \item On the other hand, if researchers do not have access to a geoparser or are interested in prioritizing area precision, we recommend using the Direct approach with a large off-the-shelf model (e.g. Llama 405B). This is due to our findings that greater precision is elicited not by providing geospatial information from an external knowledge source, but by relying on an LLM's parametric knowledge. 
\end{itemize}

\subsection{Qualitative analysis}
\label{sec:qual}

To further explore our results, we conducted a manual error analysis comparing trends in LLM outputs from the different methods. 

First, an error across model families was flipping the signs for latitude and/or longitude (e.g. "95.163" instead of "-95.163") in a predicted bounding box.
%(see examples in Table \ref{tab:error-flipped-signs}). 
We hypothesize that this error may occur due to how coordinates can represent distinctions between different hemispheres by either using positive/negative signs or cardinal directions. (e.g. "45°N" versus "45°S"). Both formats are widely used and both are likely present in LLM pretraining data corpora, so this may hinder LLMs from learning a consistent meaning for a negative or positive sign. 
We note that this error type occurred less with larger models and with the Geoparser-augmented approach, and more with Qwen versus Llama models (see Table \ref{tab:sign-flip-errors}). 
%although fine-tuning did not seem to have much of an effect. 

Next, we investigated the phenomenon identified in \S\ref{sec:results} regarding the Geoparser-augmented approach surprisingly predicting larger and less precise bounding boxes than the Direct approach. 
We observed that in LLM outputs from the Geoparser-augmented setting, the LLM would often select the maximum and minimum latitude/longitude coordinates from the mentioned locations' center coordinates and simply use these (or slightly modified versions) for the final predicted bounding box (see examples in Table \ref{tab:error-overreliance}). 
Since the mentioned locations often neighbor or contain the described location, their coordinates frequently lie outside the described location so a bounding box using these coordinates will result in a larger predicted area. 
We observed this trend across language families and model sizes, although fine-tuned and larger models seem to do more reasoning about which mentioned locations' coordinates are actually relevant for estimating the described location's bounding box. 
(LLM outputs from the Direct method typically contained only the final predicted bounding box with no step-by-step reasoning traces.)

This indicates that although providing LLMs with accurate geospatial information is generally helpful for our task, the LLMs may tend to overrely on the information without reasoning appropriately about its relationship or relevance to the desired output. 
This behavior is consistent with findings that LLMs perform much worse when irrelevant contextual information is included \cite{shi2023large} and that their ability to filter out irrelevant information decreases for more complex reasoning tasks \cite{zhou2025gsm}.
It may also be related to recent work demonstrating that reasoning models (e.g. DeepSeek-R1, Qwen3) have a strong tendency to defer to provided contextual knowledge, even when it contradicts parametric knowledge \cite{marjanovic2025deepseek}.

\section{Conclusion}

In this paper, we have introduced new methods for a challenging geocoding setting. After showing that, for our task, LLMs' geospatial reasoning skills are stronger than their geospatial knowledge, we proposed the first end-to-end methods for geocoding compositional location descriptions. Our methods achieve state-of-the-art performance in both Direct and Geoparser-augmented settings, and we offer practical options for downstream use cases. 

In addition, our novel use of bounding boxes allows us to effectively ground locations that do not exist in geographic databases and to evaluate methods with overlap-based metrics. 
Bounding boxes would likely be useful for other difficult geocoding settings, such as inherently vague geo-entities (e.g. "the Midwest") \cite{jones2008modelling}.
We encourage researchers to incorporate this way of representing locations into geocoding datasets and tools.

\newpage

\section*{Limitations}
An important limitation of this study is that we evaluate our methods on a single dataset which only includes English language examples. Since this dataset is the only existing one for our task of grounding compositional location references, we emphasize the strong need to create more datasets for this task which cover broader domains and languages. 
An additional limitation is the other possible factors which may affect performance for our task which this study does not investigate, including length of location description, number of mentioned locations in a description, frequency of an entity in pretraining data, and so on. We leave investigations of these factors on performance to future work.
Finally, we acknowledge that our proposed methods rely heavily on LLMs and are thus reliant on third parties for sustaining them.

%%%%%
%ACL 2023 requires all submissions to have a section titled “Limitations”, for discussing the limitations of the paper as a complement to the discussion of strengths in the main text. 
%This section should occur after the conclusion, but before the references. It will not count towards the page limit.

\section*{Ethical Considerations}
While the domain we evaluate here is that of formal, impersonal writing (i.e. Wikipedia), we note that it is possible to apply geocoding methods to more personal writing (e.g. social media posts) and thus risk de-anonymizing online users through the inference of sensitive location information \cite{kruspe2021changes, dupre2022geospatial}. For this reason, we have chosen to evaluate only open-source models in contrast to proprietary models, as they do not require sending potentially sensitive data to external servers and thus provide improved privacy. 
We also point to important recent work on geomasking techniques, which aim to protect the privacy of individuals while preserving spatial information in geodata \cite{lorestani2024privacy}.

%We also note the direct negative environmental impact of LLMs, with both training and inference having large energy demands \cite{rillig2023risks}. While there are arguments that LLMs can have an indirect positive environmental impact by automating tasks that would otherwise be performed by humans, it is still important to 

%%%%
%Scientific work published at ACL 2023 must comply with the ACL Ethics Policy.\footnote{\url{https://www.aclweb.org/portal/content/acl-code-ethics}} We encourage all authors to include an explicit ethics statement on the broader impact of the work, or other ethical considerations after the conclusion but before the references. The ethics statement will not count toward the page limit.

% \iffalse
\iftrue

\section*{Acknowledgements}
This material is based upon work supported by a National Science Foundation Graduate Research Fellowship (1938059) and NSF CAREER (1845576). Any opinions, findings, and conclusions or recommendations expressed in this material are those of the authors and do not necessarily reflect the views of the National Science Foundation.
\fi

\bibliography{anthology,custom}
\bibliographystyle{acl_natbib}

\newpage
\appendix

\section{Additional Experimental Details}
\label{sec:appendix-prompts}
\textbf{Data.}
For preprocessing the \textsc{GeoCoDe} dataset, we used the \emph{Beautiful Soup 4} package.\footnote{\url{https://www.crummy.com/software/BeautifulSoup/}} For accessing OSM data, we used the \emph{OSMnx 2} package \cite{boeing2025modeling}.

OSM data is available under the Open Database License.\footnote{\url{https://www.openstreetmap.org/copyright}} The license for the \textsc{GeoCoDe} dataset is unspecified,\footnote{\url{https://github.com/EgoLaparra/geocode-data}} but our use of the dataset is consistent with its intended use of benchmarking tools for geocoding compositional location references. 

\textbf{Computational experiments.}
We used \emph{together.ai}\footnote{\url{https://www.together.ai/}} for running off-the-shelf LLMs and \emph{unsloth} \cite{unsloth} for fine-tuning models. 

Our models were fine-tuned on Nvidia V100 GPUs with 16GB memory. Experiments for fine-tuning and running models required approximately 100 GPU hours in total. All results reported are from a single run (i.e. not averaged across runs).

\textbf{Prompts.}
For knowledge and reasoning baseline experiments in \S\ref{sec:disentangle}, we use Prompts \ref{fig:prompt-knowledge-point}, \ref{fig:prompt-knowledge-box}, and \ref{fig:prompt-our-approach} for LLM inference.

\begin{lstfloat}[th]
\begin{lstlisting}[breakautoindent=false, breaklines=true, breakindent=0pt]
+\textcolor{BrickRed}{SYSTEM}+: You are a system that returns the *center coordinates* of a given location or landmark. The coordinates are a pair of numbers defining the location's latitude and longitude, where latitude is a decimal number between -90.0 and 90.0 and longitude is a decimal number between -180.0 and 180.0. Follow the standard format of (latitude, longitude). Here are some examples with the expected output format:
Input: The Eiffel Tower, in France.
Output: (48.858, 2.2959)
Input: Brazil, in South America.
Output: (-14.243, -53.189)
+\textcolor{BrickRed}{USER}+: Input: +\textcolor{Fuchsia}{\textbf{\{\{location\_name\}\}}}+
Output:
\end{lstlisting}
\caption{For geospatial knowledge (point) baseline, where the input is a location name and the output is predicted center coordinates.}
\label{fig:prompt-knowledge-point}
\end{lstfloat}

\begin{lstfloat}[th]
\begin{lstlisting}[breakautoindent=false, breaklines=true, breakindent=0pt]
+\textcolor{BrickRed}{SYSTEM}+: You are a system that returns the *bounding box* of a given location or landmark. A bounding box is an area defined by two longitudes and two latitudes, where latitude is a decimal number between -90.0 and 90.0 and longitude is a decimal number between -180.0 and 180.0. Follow the standard format of (min longitude, min latitude, max longitude, max latitude). Here are some examples with the expected output format:
Input: The Eiffel Tower, in France.
Output: (2.293, 48.857, 2.297, 48.859)
Input: Brazil, in South America.
Output: (-73.983, -33.750, -34.793, 5.270)
+\textcolor{BrickRed}{USER}+: Input: +\textcolor{Fuchsia}{\textbf{\{\{location\_name\}\}}}+
Output:
\end{lstlisting}
\caption{For geospatial knowledge (box) baseline, where the input is a location name and the output is predicted bounding box.}
\label{fig:prompt-knowledge-box}
\end{lstfloat}

For approaches introduced in \S\ref{sec:geocode} for geocoding complex location references, we use Prompt \ref{fig:prompt-our-approach} for the Geoparser-augmented \texttt{reasoner} and Prompt \ref{fig:prompt-direct} for the Direct \texttt{reasoner}. 
%For the LLM-generated bounding boxes approach, we first use Prompt \ref{fig:prompt-llm-gen-bb} to generate mentioned locations' bounding boxes (i.e. function $g$) and then Prompt \ref{fig:prompt-reasoning} to generate the described location's bounding box (i.e. function $f$).
For the End-to-end LLM approach, we use Prompt \ref{fig:prompt-llm-gen-coords} for the \texttt{recaller} and Prompt \ref{fig:prompt-our-approach} for the \texttt{reasoner}.
When evaluating fine-tuned models, the examples in system prompts were excluded.

\begin{lstfloat}[th]
\begin{lstlisting}[breakautoindent=false, breaklines=true, breakindent=0pt]
+\textcolor{BrickRed}{SYSTEM}+: You are a system that returns the *bounding box* of a described location or landmark, by using a description and the center longitude and latitude of related locations. A bounding box is an area defined by two longitudes and two latitudes, where latitude is a decimal number between -90.0 and 90.0 and longitude is a decimal number between -180.0 and 180.0. Follow the standard format of (min longitude, min latitude, max longitude, max latitude). Here are some examples with the expected output format:
Input: The location is a wrought-iron lattice tower on the Champ de Mars in Paris, France. It is named after the engineer Gustave Eiffel, whose company designed and built the tower from 1887 to 1889. Champ de Mars has a longitude of 48.855 and latitude of 2.296. Paris has a longitude of 48.859 and latitude of 2.264.
Output: (2.293, 48.857, 2.297, 48.859)
Input: The location is the largest and easternmost country in South America. South America has a longitude of -13.591 and latitude of -109.712.
Output: (-73.983, -33.750, -34.793, 5.270)
+\textcolor{BrickRed}{USER}+: Input: +\textcolor{Fuchsia}{\textbf{\{\{location\_description\}\}}}+ +\textcolor{Fuchsia}{\textbf{\{\{mentioned\_location\_center\_coordinates\}\}}}+
Output:
\end{lstlisting}
\caption{For both the geospatial reasoning baseline and the Geoparser-augmented \texttt{reasoner}, where the input is a location description with mentioned locations' center coordinates and the output is predicted bounding box.}
\label{fig:prompt-our-approach}
\end{lstfloat}

\begin{lstfloat}[th]
\begin{lstlisting}[breakautoindent=false, breaklines=true, breakindent=0pt]
+\textcolor{BrickRed}{SYSTEM}+: You are a system that returns the *bounding box* of a described location or landmark. A bounding box is an area defined by two longitudes and two latitudes, where latitude is a decimal number between -90.0 and 90.0 and longitude is a decimal number between -180.0 and 180.0. Follow the standard format of (min longitude, min latitude, max longitude, max latitude). Here are some examples with the expected output format:
Input: The location is a wrought-iron lattice tower on the Champ de Mars in Paris, France. It is named after the engineer Gustave Eiffel, whose company designed and built the tower from 1887 to 1889. 
Output: (2.293, 48.857, 2.297, 48.859)
Input: The location is the largest and easternmost country in South America. 
Output: (-73.983, -33.750, -34.793, 5.270)
+\textcolor{BrickRed}{USER}+: Input: +\textcolor{Fuchsia}{\textbf{\{\{location\_description\}\}}}+ 
Output:
\end{lstlisting}
\caption{For the Direct approach \texttt{reasoner}, where the input is a location description and the output is predicted bounding box.}
\label{fig:prompt-direct}
\end{lstfloat}

\iffalse
\begin{lstfloat}[th]
\begin{lstlisting}[breakautoindent=false, breaklines=true, breakindent=0pt]
+\textcolor{BrickRed}{SYSTEM}+: You are a system that returns a list of *bounding boxes* for locations mentioned in a given paragraph. A bounding box is an area defined by two longitudes and two latitudes, where latitude is a decimal number between -90.0 and 90.0 and longitude is a decimal number between -180.0 and 180.0. Follow the standard format of (min longitude, min latitude, max longitude, max latitude). Here are some examples with the expected output format:
Input: The location is a wrought-iron lattice tower on the Champ de Mars in Paris, France. It is named after the engineer Gustave Eiffel, whose company designed and built the tower from 1887 to 1889. Return the bounding boxes for each mentioned location: [Champ de Mars, Paris].
Output: [(2.293, 48.855, 2.299, 48.859), (2.222, 48.814, 2.470, 48.904)]
Input: The location is the largest and easternmost country in South America. Return the bounding boxes for each mentioned location: [South America].
Output: [(-81.329, -53.896, -34.793, 12.459)]
+\textcolor{BrickRed}{USER}+: Input: +\textcolor{Fuchsia}{\textbf{\{\{location\_description\}\}}}+ +\textcolor{Fuchsia}{\textbf{\{\{list\_of\_mentioned\_locations\}\}}}+
Output:
\end{lstlisting}
\caption{For the LLM-generated bounding boxes approach, where the input is a location description with a list of mentioned locations and the output is mentioned locations' predicted bounding boxes.}
\label{fig:prompt-llm-gen-bb}
\end{lstfloat}
\fi

\begin{lstfloat}[th]
\begin{lstlisting}[breakautoindent=false, breaklines=true, breakindent=0pt]
+\textcolor{BrickRed}{SYSTEM}+: You are a system that returns the *center coordinates* for each location mentioned in a given paragraph. The coordinates are a pair of numbers defining each location's latitude and longitude, where latitude is a decimal number between -90.0 and 90.0 and longitude is a decimal number between -180.0 and 180.0. Here are some examples with the expected output format:
Input: The location is a wrought-iron lattice tower on the Champ de Mars in Paris, France. It is named after the engineer Gustave Eiffel, whose company designed and built the tower from 1887 to 1889. 
Output: Champ de Mars has a longitude of 48.855 and latitude of 2.296. Paris has a longitude of 48.859 and latitude of 2.264. 
Input: The location is the largest and easternmost country in South America.
Output: South America has a longitude of -13.591 and latitude of 109.712.
+\textcolor{BrickRed}{USER}+: Input: +\textcolor{Fuchsia}{\textbf{\{\{location\_description\}\}}}+ 
Output:
\end{lstlisting}
\caption{For the End-to-end LLM \texttt{recaller}, where the input is a location description and the output is mentioned locations' predicted center coordinates.}
\label{fig:prompt-llm-gen-coords}
\end{lstfloat}

\begin{table*}[t]
\centering
  \begin{tabular}{ll}
    \toprule
    \midrule\midrule
    \multicolumn{2}{l}{\parbox{16cm}{\underline{Prompt:} The location is a gulf that connects the Arabian Sea with the Strait of Hormuz, \{...\} \emph{If geo.-aug. approach:} Arabian Sea has a longitude of 63.002662154702726 and latitude of 12.4368972. Strait of Hormuz has a longitude of 56.20277021626677 and latitude of 26.449406099999997. \{...\} }} \\
    \multicolumn{2}{l}{\underline{Expected output:} \textbf{(56.2683402, 22.4824554, 61.8012822, 25.9456285)}} \\
    \midrule\midrule
    \textbf{Approach, Model} & \textbf{LLM output} \\
    \midrule
    Direct, Llama 70B & \textbf{(54.983, 22.983, 66.417, 26.750)} \\
    \midrule
    Geo.-aug., Llama 70B & \parbox{12cm}{To find the bounding box of the described location, we need to determine the minimum and maximum longitudes and latitudes. \{...\} The minimum longitude is approximately \textbf{51.1972 (Persian Gulf)} and the maximum longitude is approximately \textbf{63.0027 (Arabian Sea)} and then further to 71.2475 (Pakistan), but since the gulf is between the Arabian Sea and the Strait of Hormuz, which then runs to the Persian Gulf, the maximum longitude should be around the Arabian Sea. \{...\} So the maximum longitude should be around 63.0027.  The minimum latitude is approximately \textbf{12.4369 (Arabian Sea)} and the maximum latitude is approximately \textbf{32.6475 (Iran)}. So, the bounding box is approximately: \textbf{(51.197, 12.437, 63.003, 32.648)}} \\
    \midrule
    Direct, Llama 405B & \textbf{(58.240, 23.700, 63.320, 26.750)} \\
    \midrule
    Geo.-aug., Llama 405B & \parbox{12cm}{\{...\} - Minimum longitude: \{...\} we consider the longitude of the United Arab Emirates (53.9994829) and \textbf{the Persian Gulf (51.197231065873154)}. The minimum longitude is 51.197231065873154, but since \{...\} Therefore, we choose 51.197231065873154 as the minimum longitude, but round it to 51.2 for simplicity.  - Minimum latitude: The location is bounded by Oman on the south, so we consider the latitude of \textbf{Oman (21.0000287)}. The minimum latitude is 21.0000287, but we round it to 21.0 for simplicity.  - Maximum longitude: \{...\} we consider the longitude of \textbf{the Arabian Sea (63.002662154702726)} \{...\} Therefore, we choose 63.002662154702726 as the maximum longitude, but round it to 63.0 for simplicity.  - Maximum latitude: The location is bounded by Iran on the north, so we consider the latitude of \textbf{Iran (32.6475314)}. The maximum latitude is 32.6475314, but we round it to 32.7 for simplicity.  The bounding box of the location is \textbf{(51.2, 21.0, 63.0, 32.7)}.} \\
    \midrule
    Direct, FT Qwen 14B & \textbf{(58.0, 23.0, 62.0, 28.0)} \\
    \midrule
    Geo.-aug., FT Qwen 14B & \parbox{12cm}{\{...\} From these coordinates, the **minimum longitude** is approximately \textbf{**51.197** (Persian Gulf)}, and the **maximum longitude** is approximately \textbf{**63.003** (Arabian Sea)}. The **minimum latitude** is approximately \textbf{**21.000** (Oman)}, and the **maximum latitude** is approximately \textbf{**32.648** (Iran)}.  \#\#\# Final Answer: **\textbf{(51.197, 21.000, 63.003, 32.648)}} \\
    \bottomrule
  \end{tabular}
  \caption{Erroneous examples of LLMs overrelying on contextual geospatial information. With the Geoparser-augmented approach, LLMs tend to select bounding box coordinates directly from provided center coordinates of mentioned locations; since these center coordinates can be some distance from the actual borders of the described location, this often results in larger and less precise predictions than from the Direct approach which relies on the model's parametric knowledge.  
  For example, when used with the Geoparser-augmented approach, all three models in this table predict the Persian Gulf's longitude (51.2) as the described location's minimum longitude, much lower than its actual minimum longitude (56.3). The same three models with the Direct approach have a wider range of predictions (55.0, 58.2, 58.0) but are closer to the actual value. 
  }
  \label{tab:error-overreliance}
\end{table*}

\section{Additional Results}
\label{sec:appendix-tables}
Table \ref{tab:disentangle-complete} reports complete results from \S\ref{sec:disentangle}. Tables \ref{tab:geoparser-vs-direct-llama-complete}, \ref{tab:geoparser-vs-direct-qwen-complete}, and \ref{tab:extended-results-complete} report complete results for \S\ref{sec:results}. Table \ref{tab:error-overreliance} includes erroneous examples discussed in \S\ref{sec:qual}.

\begin{table*}[ht]
\centering
  \begin{tabular}{llrrrrr}
    \toprule
    \textbf{Experiment} & \textbf{LLM} & \textbf{Cov. (\%)} & \textbf{Dist. (km) $\downarrow$} & \textbf{AreaPrec $\uparrow$} & \textbf{AreaRec $\uparrow$} & \textbf{AreaF1 $\uparrow$}\\
    \midrule
    Knowledge (point) & Qwen 72B & 96.3 & 326.1 & -- & -- & -- \\
    & Llama 70B & 99.7 & 215.4 & -- & -- & --\\
    & Llama 405B & 97.9 & 216.6 & -- & -- & --\\
    %& Google Maps & 98.9 & 293.5 & -- & -- & -- \\
    \midrule
    Knowledge (box) & Qwen 72B & 95.6 & 333.3 & .086 & .087 & .087 \\
    & Llama 70B & \underline{99.1} & 221.8 & \underline{.199} & .105 & .137 \\
    & Llama 405B & 91.4 & 263.8 & \underline{\textbf{.361}} & .134 & .195 \\
    \midrule
    Reasoning (box) & Qwen 72B & 98.5 & \underline{\textbf{36.2}} & .191 & .368 & \underline{\textbf{.251}} \\
    & Llama 70B & 98.2 & \underline{82.9} & .166 & \underline{\textbf{.428}} & \underline{.239} \\
    & Llama 405B & \underline{\textbf{99.5}} & 92.4 & .164 & \underline{.413} & .235 \\
    \bottomrule
  \end{tabular}
  \caption{Complete evaluation metrics from our geospatial knowledge and reasoning baselines. Point refers to a center coordinate prediction; box refers to a bounding box prediction.
  The best scores are bolded, the top 2 are underlined.  } 
  \label{tab:disentangle-complete}
\end{table*}

\begin{table*}[ht]
\centering
  \begin{tabular}{llrrrrr}
    \toprule
    \textbf{Approach} & \texttt{reasoner} & \textbf{Cov. (\%)} & \textbf{Distance (km) $\downarrow$} & \textbf{AreaPrec $\uparrow$} & \textbf{AreaRec $\uparrow$} & \textbf{AreaF1 $\uparrow$}\\
    \midrule\midrule
    Direct & Llama 8B & 100.0 & 4427.8 & .000 & .000 & .000 \\
    **Geo.-aug. & Llama 8B & 25.2 & 7658.5 & .006 & .037 & .010 \\
    \midrule
    Direct & Llama 70B & 100.0 & 120.0 & .200 & .171 & .184 \\
    **Geo.-aug.. & Llama 70B & 98.2 & 82.9 & .166 & \underline{\textbf{.428}} & \underline{.239} \\
    \midrule
    Direct & Llama 405B & 99.4 & \underline{69.0} & \underline{\textbf{.305}} & .205 & \underline{\textbf{.245}} \\
    **Geo.-aug. & Llama 405B & 99.5 & 92.4 & .164 & \underline{.413} & .235 \\
    \midrule
    Direct & FT Llama 8B & 100.0 & 88.9 & .132 & .094 & .110 \\
    **Geo.-aug. & FT Llama 8B & 100.0 & \underline{\textbf{28.0}} & \underline{.237} & .186 & .208\\
    Geo.-aug. & FT Llama 8B & 100.0 & 152.2 & .231 & .170 & .196 \\
    \bottomrule
  \end{tabular}
  \caption{Results comparing Geoparser-augmented versus Direct approaches for Llama models on \textsc{GeoCoDe} dataset. FT refers to a fine-tuned model; * denotes an oracle geoparser. The best scores are bolded, the top 2 are underlined.  }
  \label{tab:geoparser-vs-direct-llama-complete}
\end{table*}

\begin{table*}[t]
\centering
  \begin{tabular}{llrrrrr}
    \toprule
    \textbf{Approach} & \texttt{reasoner} & \textbf{Cov. (\%)} & \textbf{Distance (km) $\downarrow$} & \textbf{AreaPrec $\uparrow$} & \textbf{AreaRec $\uparrow$} & \textbf{AreaF1 $\uparrow$}\\
    \midrule\midrule
    Direct & Qwen 14B & 99.1 & 1156.4 & .058 & .050 & .054 \\
    **Geo.-aug. & Qwen 14B & 100.0 & 152.1 & .134 & \underline{.498} & .211 \\
    \midrule
    Direct & Qwen 72B & 95.9 & \underline{114.7} & .113 & .162 & .133 \\  
    **Geo.-aug. & Qwen 72B & 98.5 & \underline{\textbf{36.2}} & \underline{.191} & .368 & \underline{.251} \\
    \midrule
    Direct & FT Qwen 14B & 94.7 & 1944.1 & .037 & .122 & .057 \\
    **Geo.-aug. & FT Qwen 14B & 96.1 & 135.9 & .165 & \underline{\textbf{.499}} & .248 \\
    Geo.-aug. & FT Qwen 14B & 90.8 & 240.9 & \underline{\textbf{.203}} & .384 & \underline{\textbf{.266}} \\
    \bottomrule
  \end{tabular}
  \caption{Results comparing Geoparser-augmented versus Direct approaches for Qwen models. FT refers to a fine-tuned model; * denotes an oracle geoparser. The best scores are bolded, the top 2 are underlined. }
  \label{tab:geoparser-vs-direct-qwen-complete}
\end{table*}

\begin{table*}[t]
\centering
  \begin{tabular}{llrrrrr}
    \toprule
    \textbf{Approach} & \texttt{reasoner} & \textbf{Cov. (\%)} & \textbf{Distance (km) $\downarrow$} & \textbf{AreaPrec $\uparrow$} & \textbf{AreaRec $\uparrow$} & \textbf{AreaF1 $\uparrow$}\\
    \midrule
    Geo.-aug. & FT Qwen 14B & 90.8 & \textbf{240.9} & .203 & .384 & \textbf{.266} \\
    E2E LLM & FT Qwen 14B & 95.9 & 627.5 & .100 & \textbf{.496} & .167 \\
    GBSP & Semantic parser & 52.8 & -- & \textbf{.213} & .276 & .240\\
    \bottomrule
  \end{tabular}
  \caption{Results comparing the best performing models from GBSP, End-to-end LLM, Direct, and Geoparser-augmented approaches. The best scores are bolded.}
  \label{tab:extended-results-complete}
\end{table*}

\begin{table*}[t]
\centering
  \begin{tabular}{llr}
    \toprule
    \textbf{Approach} & \textbf{Model} & \textbf{Frequency}\\
    \midrule
    Direct & Llama 70B & 2\\
    Geo.-aug. & Llama 70B & 1\\
    \midrule
    Direct & Llama 405B & 0\\
    Geo.-aug. & Llama 405B & 0\\
    \midrule
    Direct & FT Llama 8B & 0\\
    Geo.-aug. & FT Llama 8B & 0\\
    \midrule
    Direct & Qwen 14B & 15\\
    Geo.-aug. & Qwen 14B & 1\\
    \midrule
    Direct & Qwen 72B & 3\\
    Geo.-aug. & Qwen 72B & 0\\
    \midrule
    Direct & FT Qwen 14B & 3\\
    Geo.-aug. & FT Qwen 14B & 1\\
    \bottomrule
  \end{tabular}
  \caption{Flipped sign error frequencies for each model and approach from 100 random examples. We observe that the error occurs less with larger models and with the Geoparser-augmented approach. }
  \label{tab:sign-flip-errors}
\end{table*}

\end{document}